\documentclass{article}

\usepackage{microtype}
\usepackage{graphicx}
\usepackage{subfigure}
\usepackage{booktabs} 
\usepackage{amsmath}
\usepackage{amsthm}
\usepackage{amssymb}
\usepackage{mathtools}
\usepackage{xcolor}
\usepackage{csquotes}

\usepackage{placeins}

\usepackage{hyperref}
\usepackage{url}

\DeclareMathOperator*{\argmin}{arg\,min}
\newcommand\mat[1]{\mathbf{#1}}
\newcommand\mati[2]{\mat{#1}^{(#2)}}
\newcommand\matq[1]{\mat{\widehat{#1}}}
\newcommand\matsq[1]{\mat{\widetilde{#1}}}
\renewcommand\vec[1]{\mathbf{#1}}
\newcommand\veci[2]{\vec{#1}^{(#2)}}
\newcommand\vecq[1]{\vec{\widehat{#1}}}
\newcommand\dw[0]{\Delta \! \vec{w}}
\newcommand\ct[1]{\text{\rmfamily\upshape #1}}

\newcommand\func[2]{\mathnormal{#1}\left(#2\right)}
\newcommand\funcb[1]{\mathnormal{#1}}
\newcommand\tloss[1]{\mathcal{L}\left(#1\right)}
\newcommand\tlossb[0]{\mathcal{L}}
\newcommand\eop[1]{\mathop{\mathbb{E}}\left[#1\right]}

\DeclarePairedDelimiter\round{\bigg\lfloor}{\bigg\rceil}
\DeclarePairedDelimiter\ceil{\lceil}{\rceil}
\DeclarePairedDelimiter\floor{\lfloor}{\rfloor}
\DeclarePairedDelimiter\norm{\lVert}{\rVert}

\theoremstyle{definition}
\newtheorem{example}{Example}

\usepackage[accepted]{icml2020}

\icmltitlerunning{Adaptive Rounding for Post-Training Quantization}  

\begin{document}
    \newif\ifappendix
    \newif\ifcontent
    
    \contenttrue    
    \appendixtrue  
    \ifcontent
\twocolumn[
\icmltitle{Up or Down? Adaptive Rounding for Post-Training Quantization}  



\icmlsetsymbol{equal}{*}

\begin{icmlauthorlist}
\icmlauthor{Markus Nagel}{equal,qualAI}
\icmlauthor{Rana Ali Amjad}{equal,qualAI}
\icmlauthor{Mart van Baalen}{qualAI}
\icmlauthor{Christos Louizos}{qualAI}
\icmlauthor{Tijmen Blankevoort}{qualAI}
\end{icmlauthorlist}
\icmlaffiliation{qualAI}{Qualcomm AI Research, an initiative of Qualcomm Technologies, Inc.}

\icmlcorrespondingauthor{Markus Nagel}{markusn@qti.qualcomm.com}
\icmlcorrespondingauthor{Rana Ali Amjad}{ramjad@qti.qualcomm.com}
\icmlcorrespondingauthor{Tijmen Blankevoort}{tijmen@qti.qualcomm.com}

\icmlkeywords{Machine Learning, ICML, Quantization, Hessian, Rounding}

\vskip 0.3in
]



\printAffiliationsAndNotice{\icmlEqualContribution} 

\begin{abstract}
When quantizing neural networks, assigning each floating-point weight to its nearest fixed-point value is the predominant approach. We find that, perhaps surprisingly, this is not the best we can do. In this paper, we propose AdaRound, a better weight-rounding mechanism for post-training quantization that adapts to the data and the task loss. AdaRound is fast, does not require fine-tuning of the network, and only uses a small amount of unlabelled data. We start by theoretically analyzing the rounding problem for a pre-trained neural network. By approximating the task loss with a Taylor series expansion, the rounding task is posed as a quadratic unconstrained binary optimization problem. We simplify this to a layer-wise local loss and propose to optimize this loss with a soft relaxation. AdaRound not only outperforms rounding-to-nearest by a significant margin but also establishes a new state-of-the-art for post-training quantization on several networks and tasks. Without fine-tuning, we can quantize the weights of Resnet18 and Resnet50 to 4 bits while staying within an accuracy loss of 1\%.
\end{abstract}

\vspace{-0.3cm}
\section{Introduction}\label{sec:introduction}

Deep neural networks are being used in many real-world applications as the standard technique for solving tasks in computer vision, machine translation, voice recognition, ranking, and many other domains. Owing to this success and widespread applicability, making these neural networks efficient has become an important research topic. Improved efficiency translates into reduced cloud-infrastructure costs and makes it possible to run these networks on heterogeneous devices such as smartphones, internet-of-things applications, and even dedicated low-power hardware.

One effective way to optimize neural networks for inference is neural network quantization \cite{krishnamoorthi, guosurvey}. In quantization, neural network weights and activations are kept in a low-bit representation for both memory transfer and calculations in order to reduce power consumption and inference time. The process of quantizing a network generally introduces noise, which results in a loss of performance. Various prior works adapt the quantization procedure to minimize the loss in performance while going as low as possible in the number of bits used. 

As \citet{dfq} explained, the practicality of neural network quantization methods is important to take into consideration. Although many methods exist that do quantization-aware training \cite{jacob2018cvpr, louizos2018relaxed} and get excellent results, these methods require a user to spend significant time on re-training models and hyperparameter tuning. 

On the other hand, much attention has recently been dedicated to \textit{post-training quantization} methods \cite{dfq, zeroshotquant, huaweiquant, bannerposttraining}, which can be more easily applied in practice. These types of methods allow for network quantization to happen on-the-fly when deploying models, without the user of the model spending time and energy on quantization. Our work focuses on this type of network quantization.

Rounding-to-nearest is the predominant approach for all neural network weight quantization work that came out thus far. This means that the weight vector $\vec{w}$ is rounded to the nearest representable quantization grid value in a fixed-point grid by
\begin{equation}
    \widehat{\vec{w}} = \ct{s} \cdot clip \left(\round{\frac{\vec{w}}{\ct{s}}}, \ct{n}, \ct{p}\right), 
\end{equation}
 where $\ct{s}$ denotes the quantization scale parameter
and, $\ct{n}$ and $\ct{p}$ denote the negative and positive integer thresholds for clipping. We could round any weight down by replacing $\lfloor \cdot \rceil$ with $\floor{\cdot}$, or up using $\ceil{\cdot}$. But, rounding-to-nearest seems the most sensible, as it minimizes the difference per-weight in the weight matrix. Perhaps surprisingly, we show that for post-training quantization, rounding-to-nearest is not optimal. 

Our contributions in this work are threefold:
\begin{itemize}
    \item We establish a theoretical framework to analyze the effect of rounding in a way that considers the characteristics of both the input data as well as the task loss. Using this framework, we formulate rounding as a per-layer Quadratic Unconstrained Binary Optimization (QUBO) problem. 
    \item We propose AdaRound, a novel method that finds a good solution to this per-layer formulation via a continuous relaxation. AdaRound requires only a small amount of unlabelled data, is computationally efficient, and applicable to any neural network architecture with convolutional or fully-connected layers.
    \item In a comprehensive study, we show that AdaRound defines a new state-of-the-art for post-training quantization on several networks and tasks, including Resnet18, Resnet50, MobilenetV2, InceptionV3 and DeeplabV3.
\end{itemize}
\paragraph{Notation} \label{sec:notation}
We use $\vec{x}$ and $\vec{y}$ to denote the input and the target variable, respectively. $\eop{\cdot}$ denotes the expectation operator. All the expectations in this work are w.r.t. $\vec{x}$ and $\vec{y}$. $\mat{W}^{(\ell)}_{i,j}$ denotes weight matrix (or tensor as clear from the context), with the bracketed superscript and the subscript denoting the layer and the element indices, respectively. We also use $\vec{w}^{(\ell)}$ to denote flattened version of $\mat{W}^{(\ell)}$. All vectors are considered to be column vectors and represented by small bold letters, e.g., $\vec{z}$, while matrices (or tensors) are denoted by capital bold letters, e.g., $\mat{Z}$. Functions are denoted by $\mathnormal{f}(\cdot)$, except the task loss, which is denoted by $\mathcal{L}$. Constants are denoted by small upright letters, e.g., $\ct{s}$.

\section{Motivation}\label{sec:motivation}
To gain an intuitive understanding for why rounding-to-nearest may not be optimal, let's look at what happens when we perturb the weights of a pretrained model. Consider a neural network parametrized by the (flattened) weights $\vec{w}$. Let $\dw$ denote a small perturbation and $\mathcal{L(\vec{x}, \vec{y}, \vec{w})}$ denote the task loss that we want to minimize. Then
\begin{align}
&\eop{\tloss{\vec{x}, \vec{y}, \vec{w} + \dw} - \tloss{\vec{x}, \vec{y}, \vec{w}}} \\
\overset{(a)}{\approx} & \mathbb{E}\left[\dw^T \cdot \nabla_{\vec{w}} \tloss{\vec{x}, \vec{y}, \vec{w}}\right. \nonumber \\
&\qquad + \left. \frac{1}{2}\dw^T \cdot \nabla^2_{\vec{w}} \tloss{\vec{x}, \vec{y}, \vec{w}} \cdot \dw \right]  \\
=  & \quad  \dw^T \cdot \veci{g}{\vec{w}} + \frac{1}{2}\dw^T \cdot \mati{H}{\vec{w}} \cdot \dw , 
\label{eqn:lossfunc}
\end{align}
where (a) uses the second order Taylor series expansion. $\veci{g}{\vec{w}}$ and $\mati{H}{\vec{w}}$ denote the expected gradient and Hessian of the task loss $\tlossb$ w.r.t. $\vec{w}$, i.e.,
\begin{align}
    \veci{g}{\vec{w}} &= \eop{\nabla_{\vec{w}} \tloss{\vec{x}, \vec{y}, \vec{w}}}  \\
    \mati{H}{\vec{w}} &= \eop{\nabla^2_{\vec{w}} \mathcal{L}(\vec{x}, \vec{y}, \vec{w})}. 
\end{align}
All the gradient and Hessian terms in this paper are of task loss $\tlossb$ with respect to the specified variables. Ignoring the higher order terms in the Taylor series expansion is a good approximation as long as $\dw$ is not too large. Assuming the network is trained to convergence, we can also ignore the gradient term as it will be close to $0$. Therefore, $\mati{H}{\vec{w}}$ defines the interactions between different perturbed weights in terms of their joint impact on the task loss $\tloss{\vec{x},\vec{y},\vec{w}+\dw}$. The following toy example illustrates how rounding-to-nearest may not be optimal. 
\begin{example} \label{exp:toyexp}
Assume $\dw^T = [\Delta w_1 \quad \Delta w_2]$ and  
\begin{align}
    \mati{H}{\vec{w}} = \begin{bmatrix}
           1 & 0.5 \\           
           0.5 & 1
    \end{bmatrix}, 
\end{align}
then the increase in task loss due to the perturbation is (approximately) proportional to \vspace{-2pt}
\begin{align}
    \dw^T \cdot \mati{H}{\vec{w}} \cdot \dw &=  \Delta \vec{w}_1^2 + \Delta \vec{w}_2^2 + \Delta \vec{w}_1 \Delta \vec{w}_2. 
\end{align}
For the terms corresponding to the diagonal entries $\Delta \vec{w}_1^2$ and $\Delta \vec{w}_2^2$, only the magnitude of the perturbations matters. Hence rounding-to-nearest is optimal when we only consider these diagonal terms in this example. However, for the terms corresponding to the $\Delta \vec{w}_1 \Delta \vec{w}_2$, the sign of the perturbation matters, where opposite signs of the two perturbations improve the loss. To minimize the overall impact of quantization on the task loss, we need to trade-off between the contribution of the diagonal terms and the off-diagonal terms. Rounding-to-nearest ignores the off-diagonal contributions, making it often sub-optimal.
\end{example}
The previous analysis is valid for the quantization of any parametric system. We show that this effect also holds for neural networks. To illustrate this, we generate 100 stochastic rounding \cite{Gupta2015} choices for the first layer of Resnet18 and evaluate the performance of the network with only the first layer quantized. The results are presented in Table~\ref{tbl:motivation}. Among $100$ runs, we find that $48$ stochastically sampled rounding choices lead to a better performance than rounding-to-nearest. This implies that many rounding solutions exist that are better than rounding-to-nearest. Furthermore, the best among these 100 stochastic samples provides more than $10\%$ improvement in the accuracy of the network. We also see that accidentally rounding all values up, or all down, has an catastrophic effect. This implies that we can gain a lot by carefully rounding weights when doing post-training quantization. The rest of this paper is aimed at devising a well-founded and computationally efficient rounding mechanism.

\begin{table}[t]
    \centering
    \begin{tabular}{ l r }
        \toprule
         Rounding scheme          & Acc(\%)       \\\midrule
         Nearest            & 52.29  \\
         Ceil               & 0.10   \\
         Floor              & 0.10  \\\midrule
         Stochastic         & 52.06$\pm$5.52   \\ 
         Stochastic (best)  & 63.06  \\
         \bottomrule  
    \end{tabular}\vspace{-.1cm}
    \caption{Comparison of ImageNet validation accuracy among different rounding schemes for $4$-bit quantization of the first layer of Resnet18. We report the mean and the standard deviation of 100 stochastic \cite{Gupta2015} rounding choices (Stochastic) as well as the best validation performance among these samples (Stochastic (best)).}
    \label{tbl:motivation}\vspace{-.1cm}
\end{table}

\section{Method}\label{sec:method}
In this section, we propose AdaRound, a new rounding procedure for post-training quantization that is theoretically well-founded and shows significant performance improvement in practice. We start by analyzing the loss due to quantization theoretically. We then formulate an efficient per-layer algorithm to optimize it.

\subsection{Task loss based rounding}\label{sec:otr}
When quantizing a pretrained NN, our aim is to minimize the performance loss incurred due to quantization. Assuming per-layer weight quantization\footnote{Note that our work is equally applicable for per-channel weight quantization.}, the quantized weight $\widehat{\vec{w}}_i^{(\ell)}$ is
\begin{align}
    \widehat{\vec{w}}_i^{(\ell)}  \in \left\{\vec{w}_i^{(\ell), floor}, \vec{w}_i^{(\ell), ceil}\right\}, \label{eq:wquant}
\end{align} 
where 
\begin{align}
    \vec{w}_i^{(\ell), floor} &=  \ct{s}^{(\ell)} \cdot clip\left(\floor*{\frac{\vec{w}_i^{(\ell)}}{\ct{s}^{(\ell)}}}, \ct{n} , \ct{p} \right) \label{eq:floor}
\end{align}
and $\vec{w}_i^{(\ell), ceil}$ is similarly defined by replacing $\floor*{\cdot}$ with $\ceil*{\cdot}$ and $\dw_i^{(\ell)} = \veci{w}{\ell} - \widehat{\vec{w}}_i^{(\ell)}$ denotes the perturbation due to quantization. In this work we assume $\ct{s}^{(\ell)}$ to be fixed prior to optimizing the rounding procedure. Finally, whenever we optimize a cost function over the $\dw_i^{(\ell)}$, the $\widehat{\vec{w}}_i^{(\ell)}$ can only take two values specified in \eqref{eq:wquant}. 

Finding the optimal rounding procedure can be formulated as the following binary optimization problem 
\begin{align}
    \argmin_{\dw}& \quad \eop{\tloss{\vec{x},\vec{y},\vec{w} + \dw} - \tloss{\vec{x},\vec{y},\vec{w}}} \label{eq:loss_opt}
\end{align}
Evaluating the cost in \eqref{eq:loss_opt} requires a  forward pass of the input data samples for each new $\dw$ during optimization. To avoid the computational overhead of repeated forward passes throught the data, we utilize the second order Taylor series approximation. Additionally, we ignore the interactions among weights belonging to different layers. This, in turn, implies that we assume a block diagonal $\mati{H}{\vec{w}}$, where each non-zero block corresponds to one layer. We thus end up with the following per-layer optimization problem
\begin{align}
    \argmin_{\dw^{(\ell)}}& \quad \eop{ {\vec{g}^{(\vec{w}^{(\ell)})}}^T \dw^{(\ell)} + \frac{1}{2}{\dw^{(\ell)}}^T \mati{H}{\veci{w}{\ell}} \dw^{(\ell)}}. \label{eq:taylor_opt0}
\end{align}
As illustrated in Example~\ref{exp:toyexp}, we require the second order term to exploit the joint interactions among the weight perturbations. \eqref{eq:taylor_opt0} is a QUBO problem since $\dw_i^{(\ell)}$ are binary variables \cite{Kochenberger2014}. For a converged pretrained model, the contribution of the gradient term for optimization in \eqref{eq:taylor_opt} can be safely ignored. This results in
\begin{align}
    \argmin_{\dw^{(\ell)}}& \quad \eop{{\dw^{(\ell)}}^T \mati{H}{\veci{w}{\ell}} \dw^{(\ell)}}. \label{eq:taylor_opt}
\end{align}
\begin{figure}[t]
\centering
  \includegraphics[width=\linewidth -.5cm]{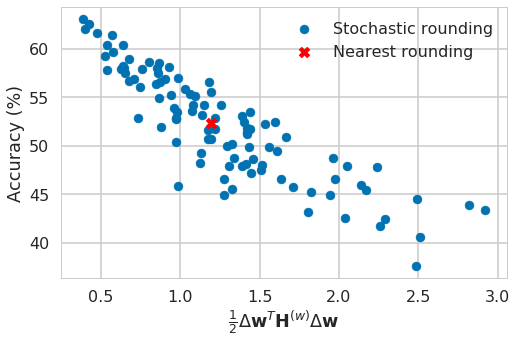}\vspace{-0.3cm}
 \caption{Correlation between the cost in \eqref{eq:taylor_opt} vs ImageNet validation accuracy (\%) of 100 stochastic rounding vectors $\vecq{w}$ for $4$-bit quantization of only the first layer of Resnet18.}\vspace{-0.1cm}
 \label{fig:cost_vs_acc}
\end{figure}
To verify that \eqref{eq:taylor_opt} serves as a good proxy for optimizing task loss due to quantization,  we plot the cost in \eqref{eq:taylor_opt} vs validation accuracy for 100 stochastic rounding vectors when quantizing only the first layer of Resnet18. Fig.~\ref{fig:cost_vs_acc} shows a clear correlation between the two quantities. This justifies our approximation for optimization, even for $4$ bit quantization. 
Optimizing \eqref{eq:taylor_opt} show significant performance gains, however its application is limited by two problems:
\begin{enumerate}
    \item $\mati{H}{\veci{w}{\ell}}$ suffers from both computational as well memory complexity
    issues even for moderately sized layers.
    \item \eqref{eq:taylor_opt} is an NP-hard optimization problem. The complexity of solving  it scales rapidly with the dimension of $\dw^{(\ell)}$, again prohibiting the application of \eqref{eq:taylor_opt} to even moderately sized layers \cite{Kochenberger2014}.
\end{enumerate}
In section~\ref{sec:localloss} and section~\ref{sec:ltr} we tackle the first and the second problem, respectively.

\subsection{From Taylor expansion to local loss}\label{sec:localloss}
To understand the cause of the complexity associated with $\mati{H}{\veci{w}{\ell}}$, let us look at its' elements. For two weights in the same fully connected layer we have 
\begin{align}
    \frac{\partial^2 \tlossb}{\partial \mati{W}{\ell}_{i,j} \partial \mati{W}{\ell}_{m,o}} &= \frac{\partial }{\partial \mati{W}{\ell}_{m,o}} \left[\frac{\partial \tlossb}{\partial \veci{z}{\ell}_i} \cdot \veci{x}{\ell-1}_j\right]  \\
    &= \frac{\partial^2 \tlossb}{\partial \veci{z}{\ell}_i \partial \veci{z}{\ell}_m} \cdot \veci{x}{\ell-1}_j \veci{x}{\ell-1}_o,
\end{align}
where $\veci{z}{\ell} = \mati{W}{\ell}\veci{x}{\ell-1}$ are the preactivations for layer $\ell$ and $\veci{x}{\ell-1}$ denotes the input to layer $\ell$. Writing this in matrix formulation (for flattened $\veci{w}{\ell}$), we have \cite{gauss-newton2017}
\begin{align}
  \mati{H}{\mati{w}{\ell}}  &= \eop{\veci{x}{\ell-1}{\veci{x}{\ell-1}}^{T} \otimes \nabla^2_{\veci{z}{\ell}} \tlossb }, \label{eq:fchesskron0}
\end{align}
where $\otimes$ denotes Kronecker product of two matrices and $\nabla^2_{\veci{z}{\ell}} \tlossb$ is the Hessian of the task loss w.r.t. $\veci{z}{\ell}$. It is clear from \eqref{eq:fchesskron0} that the complexity issues are mainly caused by $\nabla^2_{\veci{z}{\ell}} \tlossb$ that requires backpropagation of second derivatives through the subsequent layers of the network. To tackle this, we make the assumption that the Hessian of the task loss w.r.t. the preactivations, i.e., $\nabla^2_{\veci{z}{\ell}} \tlossb$ is a diagonal matrix, denoted by $diag\left(\nabla^2_{\veci{z}{\ell}} \tlossb_{i,i}\right)$. This leads to 
\begin{align}
  \mati{H}{\mati{w}{\ell}}  &= \eop{\veci{x}{\ell-1}{\veci{x}{\ell-1}}^T \otimes diag(\nabla^2_{\veci{z}{\ell}} \tlossb_{i,i})}. \label{eq:fchesskron}
\end{align}
Note that the approximation of $\mati{H}{\mati{w}{\ell}}$  expressed in \eqref{eq:fchesskron} is not diagonal.  Plugging \eqref{eq:fchesskron} into our equation for finding the rounding vector that optimizes the loss \eqref{eq:taylor_opt}, we obtain
\small
\begin{align}
     &\argmin_{\Delta \mati{W}{\ell}_{k,:}} \quad \eop{\nabla^2_{\veci{z}{\ell}} \tlossb_{k,k} \cdot \Delta \mati{W}{\ell}_{k,:}  \veci{x}{\ell-1}{\veci{x}{\ell-1}}^T \Delta {\mati{W}{\ell}_{k,:}}^T} \label{eq:subhessopt0} \\
     &\overset{(a)}{=} \argmin_{\Delta \mati{W}{\ell}_{k,:}} \quad \Delta \mati{W}{\ell}_{k,:}  \eop{\veci{x}{\ell-1}{\veci{x}{\ell-1}}^T} \Delta {\mati{W}{\ell}_{k,:}}^T \label{eq:subhessopt1} \\
      &=  \argmin_{\Delta \mati{W}{\ell}_{k,:}} \quad \eop{\left(\Delta \mati{W}{\ell}_{k,:}
      \veci{x}{\ell-1}\right)^2}, \label{eq:subhessopt2}
\end{align}
\normalsize
where the optimization problem in \eqref{eq:taylor_opt} now decomposes into independent sub-problems in \eqref{eq:subhessopt0}. Each sub-problem deals with a single row $\Delta \mati{W}{\ell}_{k,:}$  and (a) is the outcome of making a further assumption that $\nabla^2_{\veci{z}{\ell}} \tlossb_{i,i} =\ct{c}_i$ is a constant independent of the input data samples. It is worthwhile to note that optimizing \eqref{eq:subhessopt2} requires no knowledge of the subsequent layers and the task loss. In \eqref{eq:subhessopt2}, we are simply minimizing the Mean Squared Error (MSE) introduced in the preactivations $\veci{z}{\ell}$ due to quantization. 
This is the same layer-wise objective that was optimized in several neural network compression papers, e.g., \citet{zhang2015, he2017}, and various neural network quantization papers (albeit for tasks other than weight rounding), e.g., \citet{TSQ2018, BGD2020, huaweiquant}. However, unlike these works, we arrive at this objective in a principled way and conclude that optimizing the MSE, as specified in \eqref{eq:subhessopt2}, is the best we can do when assuming no knowledge of the rest of the network past the layer that we are optimizing.  In the supplementary material we perform an analogous analysis for convolutional layers.

The optimization problem in \eqref{eq:subhessopt2} can be tackled by either precomputing $\eop{\veci{x}{\ell-1}{\veci{x}{\ell-1}}^ T}$, as done in \eqref{eq:subhessopt1}, and then performing the optimization over $\Delta \mati{W}{\ell}_{k,:}$, or by performing a single layer forward pass for each potential $\Delta \mati{W}{\ell}_{k,:}$ during the optimization procedure.

 In section~\ref{sec:experiments}, we empirically verify that the constant diagonal approximation of $\nabla^2_{\veci{z}{\ell}} \tlossb$ does not negatively influence the performance.

\subsection{AdaRound}\label{sec:ltr}
Solving \eqref{eq:subhessopt2} does not suffer from complexity issues associated with $\mati{H}{\veci{w}{\ell}}$. However, it is still an NP-hard discrete optimization problem. Finding good (sub-optimal) solution with reasonable computational complexity can be a challenge for larger number of optimization variables. To tackle this we relax \eqref{eq:subhessopt2} to the following continuous optimization problem based on soft quantization variables (the superscripts are the same as \eqref{eq:subhessopt2})
\begin{equation}
    \argmin_{\mat{V}} \quad \norm*{\mat{W} \vec{x} - \matsq{W} \vec{x}}^2_F + \lambda \func{f_{reg}}{\mat{V}}, \label{eq:mse_relax}
\end{equation}
where $\norm*{\cdot}^2_F$ denotes the Frobenius norm and $\matsq{W}$ are the soft-quantized weights that we optimize over 
\begin{align}
    \matsq{W} &= \ct{s} \cdot clip \left(\floor*{\frac{\mat{W}}{\ct{s}}} + \func{h}{\mat{V}}, \ct{n}, \ct{p} \right).  \label{eq:soft_quant}
\end{align}
In the case of a convolutional layer the $\mat{W} \vec{x}$ matrix multiplication is replaced by a convolution. $\mat{V}_{i,j}$ is the continuous variable that we optimize over and $\func{h}{\mat{V}_{i,j}}$ can be any differentiable function that takes values between $0$ and $1$, i.e.,  $\func{h}{\mat{V}_{i,j}} \in [0, 1]$. The additional term $\func{f_{reg}}{\mat{V}}$ is a differentiable regularizer that is introduced to encourage the optimization variables $\func{h}{\mat{V}_{i,j}}$ to converge towards either $0$ or $1$, i.e., at convergence $\func{h}{\mat{V}_{i,j}} \in \{0, 1\}$. 

We employ a rectified sigmoid as $\func{h}{\mat{V}_{i,j}}$, proposed in \cite{louizos2018learning}. The rectified sigmoid is defined as  
\begin{align}
    \func{h}{\mat{V}_{i,j}} &=  clip(\sigma \left(\mat{V}_{i,j} \right) (\zeta - \gamma) + \gamma, 0, 1), \label{eq:rect_sigmoid}
\end{align}
where $\sigma(\cdot)$ is the sigmoid function and, $\zeta$ and $\gamma$ are stretch parameters, fixed to $1.1$ and $-0.1$, respectively. The rectified sigmoid has non-vanishing gradients as $\func{h}{\mat{V}_{i,j}}$ approaches $0$ or $1$, which helps the learning process when we encourage $\func{h}{\mat{V}_{i,j}}$ to move to the extremities. For regularization we use 
\begin{equation}
    \func{f_{reg}}{\mat{V}} = \sum\limits_{i,j} 1 - |2 \func{h}{\mat{V}_{i,j}} - 1|^\beta, \label{eq:reg_func}
\end{equation}
where we anneal the parameter $\beta$. This allows most of the $\func{h}{\mat{V}_{i,j}}$ to adapt freely in the initial phase (higher $\beta$) to improve the MSE and encourages it to converge to $0$ or $1$ in the later phase of the optimization (lower $\beta$), to arrive at the binary solution that we are interested in. The effect of annealing $\beta$ is illustrated in Fig.~\ref{fig:reg_loss}. Fig.~\ref{fig:movement} shows how this combination of rectified sigmoid and $ \funcb{f_{reg}}$ leads to many weights learning a rounding that is different from rounding to the nearest, to improve the performance, while ultimately converging close to $0$ or $1$. 

This method of optimizing \eqref{eq:mse_relax} is a specific instance of the general family of Hopfield methods used for binary constrained optimization problems. These types of methods are commonly used as an efficient approximation algorithm for large scale combinatorial problems \cite{Hopfield1985, smithhopfield}.

\begin{figure}[t]
    \includegraphics[width=\linewidth -.7cm, trim={.2cm .2cm .2cm .2cm}, clip]{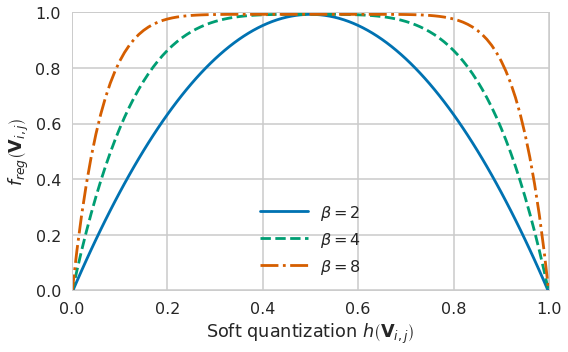}\vspace{-0.3cm}
    \centering
    \caption{Effect of annealing $b$ on regularization term \eqref{eq:reg_func}.}\vspace{-0.3cm}
\label{fig:reg_loss}
\end{figure}

To quantize the whole model, we optimize \eqref{eq:mse_relax} layer-by-layer sequentially. However, this does not account for the quantization error introduced due to the previous layers. In order to avoid the accumulation of quantization error for deeper networks as well as to account for the activation function, we use the following asymmetric reconstruction formulation 
\begin{equation}
    \argmin_{\mat{V}} \norm*{ \func{f_{a}}{\mat{W} \vec{x}} - \func{f_{a}}{\matsq{W} \vec{\hat{x}}}}^2_F + \lambda \func{f_{reg}}{\mat{V}}, \label{eq:mse_asym}
\end{equation}
where $\vec{\hat{x}}$ is the layer's input with all preceding layers quantized and $\funcb{f_a}$ is the activation function. A similar formulation of the loss has been used previously in \cite{zhang2015, he2017}, albeit for different purposes. \eqref{eq:mse_asym} defines our final objective that we can optimize via stochastic gradient descent. We call this algorithm AdaRound, as it adapts to the statistics of the input data as well as to (an approximation of) the task loss. In section~\ref{sec:experiments} we elaborate on the influence of our design choices as well as the asymmetric reconstruction loss on the performance.  

\begin{figure}[t]
    \includegraphics[width=\linewidth -.7cm, trim={.2cm .2cm .2cm .2cm}, clip]{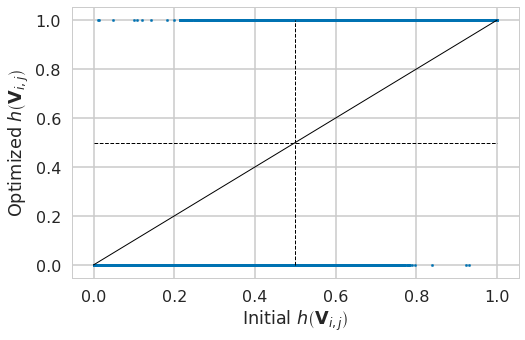}\vspace{-0.3cm}
    \centering
    \caption{Comparison of $\func{h}{\mat{V}_{i,j}}$ before (x-axis, corresponding to floating point weights) vs after (y-axis) optimizing \eqref{eq:mse_relax}. We see that all   $\func{h}{\mat{V}_{i,j}}$ have converged to $0$ or $1$. Top left and lower right quadrants indicate the weights that have different rounding using \eqref{eq:mse_relax} vs rounding-to-nearest.}\vspace{-0.3cm}
\label{fig:movement}
\end{figure}

\section{Background and related work}\label{sec:backgroundrelated}
In the 1990s, with the resurgence of the field of neural networks, several works designed hardware and optimization methods for running low-bit neural networks on-device. \citet{hammerstrom} created hardware for 8 and 16-bit training of networks, \citet{holihwang} did an empirical analysis on simple neural networks to show that 8 bits are sufficient in most scenarios, and \citet{oldstochasticrounding} developed a stochastic rounding scheme to push neural networks below 8 bits.

More recently, much attention has gone to quantizing neural networks for efficient inference. This is often done by simulating quantization during training, as described in \citet{jacob2018cvpr} and \citet{Gupta2015}, and using a straight-through estimator to approximate the gradients. Many methods have since then extended these training frameworks. \citet{pact2018} learns the activations to obey a certain quantization range, while \citet{lsq, xilinxquant} learn the quantization min and max ranges during training so that they do not have to be set manually. \citet{louizos2018relaxed} also learn the grid and formulate a probabilistic version of the quantization training procedure. \citet{differentiablequantization} learn both the quantization grid, and the bit-width per layer, resulting in automatic bit-width selection during training. Works like \citet{jangho, apprentice} exploit student-teacher training to improve quantized model performance during training. Although quantization-aware training is potent and often gives good results, the process is often tedious and time-consuming. Our work seeks to get high accuracy models without this hassle.

Several easy-to-use methods for quantization of networks without quantization-aware training have been proposed as of recent. These methods are often referred to as \textit{post-training quantization} methods. \citet{krishnamoorthi} show several results of network quantization without fine-tuning. Works like \citet{bannerposttraining, huaweiquant} optimize the quantization ranges for clipping to find a better loss trade-off per-layer. \citet{OCS} improve quantization performance by splitting channels into more channels, increasing computation but achieving lower bit-widths in the process. \citet{lin2016icml, hawq} set different bit-widths for different layers, through the information of the per-layer SQNR or the Hessian. \citet{dfq, zeroshotquant} even do away with the requirement of needing any data to optimize a model for quantization, making their procedures virtually parameter and data-free. These methods are all solving the same quantization problem as in this paper, and some like \citet{OCS} and \citet{hawq} could even be used in conjunction with AdaRound. We compare to the methods that improve weight quantization for 4/8 and 4/32 bit-widths without end-to-end fine-tuning, \citet{bannerposttraining, huaweiquant, dfq}, but leave out comparisons to the mixed-precision methods \citet{zeroshotquant, hawq} since they improve networks on a different axis. 

\section{Experiments}\label{sec:experiments}
To evaluate the performance of AdaRound, we conduct experiments on various computer vision tasks and models. In section~\ref{sec:ablation} we study the impact of the approximations and design choices made in section~\ref{sec:method}. In section~\ref{sec:comp_literature} we compare AdaRound to other post-training quantization methods.

\paragraph{Experimental setup}
For all experiments we absorb batch normalization in the weights of the adjacent layers.
We use symmetric $4$-bit weight quantization with a per-layer scale parameter $\ct{s}^{(\ell)}$
which is determined prior to the application of AdaRound. 
We set $\ct{s}$ so that it minimizes the MSE $||\mat{W} - \overline{\mat{W}}||^2_F$, where $\overline{\mat{W}}$ are the quantized weights obtained through rounding-to-nearest.
In some ablation studies, we report results when quantizing only the first layer. This will be explicitly mentioned as \enquote{First layer}. In all other cases, we have the weights of the whole network quantized using $4$ bits. Unless otherwise stated, all activations are in FP32.
Most experiments are conducted using Resnet18 \cite{heresidual} from torchvision. The baseline performance of this model with full precision weights and activations is 69.68\%. In our experiments, we report the mean and standard deviation of the (top1) accuracy on the ImageNet validation set, calculated using 5 runs with different initial seeds.
To optimize AdaRound we use 1024 unlabeled images from the ImageNet \cite{imagenet} training set, Adam \cite{kingma2014adam} optimizer with default hyper-parameters for 10k iterations and a batch-size of 32, unless otherwise stated. We use Pytorch \cite{pytorch} for all our experiments.
It is worthwhile to note that the application of AdaRound  to Resnet18 takes only 10 minutes on a single Nvidia GTX 1080 Ti.

\subsection{Ablation study}\label{sec:ablation} 

\paragraph{From task loss to local loss}
We make various approximations and assumptions in section~\ref{sec:otr} and section~\ref{sec:localloss} to simplify our optimization problem. In Table~\ref{tbl:exp2_hessian}, we look at their impact systematically. First, we note that optimizing based on the Hessian of the task loss (cf. \eqref{eq:taylor_opt}) provides a significant performance boost compared to rounding-to-nearest. This verifies that the Taylor expansion based rounding serves as a much better alternative for the task loss when compared to rounding-to-nearest. Similarly, we show that, although moving from the optimization of Taylor expansion of the task loss to the local MSE loss (cf. \eqref{eq:subhessopt2}) requires strong assumptions, it does not degrade the performance. Unlike the Taylor series expansion, the local MSE loss makes it feasible to optimize all layers in the network. We use the cross-entropy method \cite{Rubinstein1999} to solve the QUBO problems in \eqref{eq:taylor_opt} and \eqref{eq:subhessopt2}, where we initialize the sampling distribution for the binary random variables $\widehat{\vec{w}}_i$ as in \cite{Gupta2015}\footnote{In the supplementary material we compare the performance of different QUBO solvers on our problem.}. Finally, the continuous relaxation for the local MSE optimization problem (cf. \eqref{eq:mse_relax})
not only reduces the optimization time from several hours to a few minutes but also slightly improves our performance.

\begin{table}[t]
    \centering
    \begin{tabular}{ l r r }
        \toprule
         Rounding          & First layer      & All layers \\\midrule
         Nearest            &  52.29   &  23.99 \\
         $\mati{H}{\vec{w}}$ task loss (cf. \eqref{eq:taylor_opt}) & 68.62$\pm$0.17   &  N/A \\
         Local MSE loss (cf. \eqref{eq:subhessopt2}) &  69.39$\pm$0.04  &  65.83$\pm$0.14 \\ 
         Cont. relaxation (cf \eqref{eq:mse_relax})     &  \textbf{69.58$\pm$0.03}   &  \textbf{66.56$\pm$0.12} \\
         \bottomrule
    \end{tabular}\vspace{-0.1cm}
    \caption{Impact of various approximations and assumptions made in section~\ref{sec:method} on the ImageNet validation accuracy (\%) for Resnet18. N/A implies that the corresponding experiment was computationally infeasible.}\vspace{-0.1cm}
    \label{tbl:exp2_hessian}
\end{table}

\paragraph{Design choices for AdaRound}
As discussed earlier, our approach to solve \eqref{eq:mse_relax} closely resembles a Hopfield method. These methods optimize $\func{h}{\mat{V}_{i,j}} = \sigma\left(\frac{\mat{V}_{i,j}}{T}\right)$ with a version of gradient descent with respect to $V_{i,j}$, and annealing the temperature $T$ \cite{Hopfield1985, smithhopfield}. This annealing acts as an implicit regularization that allows $\func{h}{\mat{V}_{i,j}}$ to optimize for the MSE loss initially unconstrained, while encouraging $\func{h}{\mat{V}_{i,j}}$ to converge towards $0$ or $1$ in the later phase of optimization. In Table \ref{tbl:exp3_reg} we show that even after an extensive hyper-parameter search for the annealing schedule of $T$, using the sigmoid function with our explicit regularization term \eqref{eq:reg_func} outperforms the classical method. Using explicit regularization also makes the optimization more stable, leading to lower variance as shown in Table~\ref{tbl:exp3_reg}. Furthermore, we see that the use of the rectified sigmoid also provides a consistent small improvement in accuracy for different models.

Table \ref{tbl:exp4_asym_relu} shows the gain of using the asymmetric reconstruction MSE (cf. section~\ref{sec:ltr}). We see that this provides a noticeable accuracy improvement when compared to \eqref{eq:mse_relax}. Similarly, accounting for the activation function in the optimization problem provides a small gain.

\begin{table}[t]
    \centering
    \begin{tabular}{ l r r }
        \toprule
         Rounding                           & First layer      & All layers \\\midrule
         Sigmoid + $T$ annealing               & 69.31$\pm$0.21  &  65.22$\pm$0.67 \\
         Sigmoid + $\funcb{f_{reg}}$                & \textbf{69.58$\pm$0.03}   &  66.25$\pm$0.15   \\ 
         Rect. sigmoid + $\funcb{f_{reg}}$      &  \textbf{69.58$\pm$0.03}   &  \textbf{66.56$\pm$0.12} \\  
         \bottomrule
    \end{tabular}\vspace{-0.1cm}
    \caption{Impact of different design choices for optimizing \eqref{eq:mse_relax}, on  the  ImageNet  validation accuracy (\%) for Resnet18.}\vspace{-0.1cm}
    \label{tbl:exp3_reg}
\end{table}
\begin{table}[t]
    \centering
    \begin{tabular}{ l r }
        \toprule
         Optimization           & Acc (\%)   \\\midrule
         Layer wise             & 66.56$\pm$0.12  \\
         Asymmetric             & 68.37$\pm$0.07 \\
         Asymmetric + ReLU      & \textbf{68.60$\pm$0.09} \\
         \bottomrule 
    \end{tabular}\vspace{-0.1cm}
    \caption{The influence on the ImageNet validation accuracy (\%) for Resnet18, by incorporating asymmetric reconstruction MSE loss and activation function in the rounding optimization objective.}
    \label{tbl:exp4_asym_relu}\vspace{-0.1cm}
\end{table}
  
\paragraph{Optimization using STE}
Another option we considered is to optimize $\matsq{W}$ directly by using the straight-through estimator (STE) \cite{bengio2013estimating}. This is inspired by quantization-aware training \cite{jacob2018cvpr}, which optimizes a full network with this procedure. We use the STE to minimize the MSE loss in \eqref{eq:mse_relax}. This method technically allows more flexible movement of the quantized weights $\matq{W}$, as they are no longer restricted to just rounding up or down. 
In Table \ref{tbl:exp5_ste} we compare the STE optimization with AdaRound. We can see that AdaRound clearly outperforms STE-based optimization. We believe this is due to the biased gradients of the STE, which hinder the optimization in this restricted setting. 

\begin{table}[t]
    \centering
    \begin{tabular}{ l r }
        \toprule
         Optimization           & Acc (\%)   \\\midrule
         Nearest            & 23.99  \\
         STE                & 66.63$\pm$0.06 \\
         AdaRound           & \textbf{68.60$\pm$0.09} \\
         \bottomrule 
    \end{tabular}\vspace{-0.1cm}
    \caption{Comparison between optimizing \eqref{eq:mse_asym} using STE (without explicit regularization $\funcb{f_{reg}}$) vs AdaRound. We report ImageNet validation accuracy (\%) for Resnet18.}
    \label{tbl:exp5_ste}\vspace{-0.1cm}
\end{table}

\paragraph{Influence of quantization grid}\label{subsec:grid}
 We studied how the choice of weight quantization grid affects the performance gain that AdaRound brings vs rounding-to-nearest. We looked at three different options for determining the scale parameter $\ct{s}$; using minimum and maximum values of the weight tensor $\mat{W}$, minimizing the MSE $\norm*{\mat{W} - \overline{\mat{W}}}^2_F$ introduced in the weights, and minimizing the MSE $\norm*{\mat{W}\vec{x} - \overline{\mat{W}}\vecq{x}}^2_F$ introduced in the preactivations. $\overline{\mat{W}}$ denotes the quantized weight tensor obtained through rounding-to-nearest for a given $\ct{s}$.  Note, we do not optimize step size and AdaRound jointly as it is non-trivial to combine the two tasks: any change in the step size would result in a different QUBO problem.  The results in Table \ref{tbl:exp51_grid} clearly show that AdaRound significantly improves over rounding-to-nearest, independent of the choice of the quantization grid. Both MSE based approaches are superior to the Min-Max method for determining the grid. Since there is no clear winner between the two MSE formulations for AdaRound, we continue the use of $\norm*{\mat{W} - \overline{\mat{W}}}^2_F$  formulation for all other experiments.

\begin{table}[t]
    \centering
    \begin{tabular}{ l r r }
        \toprule
         Grid                                                   & Nearest       & AdaRound \\\midrule
         Min-Max                                                 & 0.23          & 61.96$\pm$0.04 \\
        $\norm*{\mat{W} - \overline{\mat{W}}}^2_F$             & 23.99         & \textbf{68.60$\pm$0.09} \\
        $\norm*{\mat{W}\vec{x} - \overline{\mat{W}}\vecq{x}}^2_F$   & 42.89         & \textbf{68.62$\pm$0.08} \\
         \bottomrule 
    \end{tabular}\vspace{-0.1cm}
    \caption{Comparison between various quantization grids in combination with rounding-to-nearest and AdaRound. We report ImageNet validation accuracy (\%) for Resnet18.}
    \label{tbl:exp51_grid}\vspace{-0.1cm}
\end{table}


\paragraph{Optimization robustness to data}
We also investigate how little data is necessary to allow AdaRound to achieve good performance and investigate if this could be done with data from different datasets. The results can be seen in Fig.~\ref{fig:datasets}. We see that the performance of AdaRound is robust to the number of images required for optimization. Even with as little as $256$ images, the method optimizes the model to within $2\%$ of the original FP32 accuracy. We also see that when using unlabelled images that are from a similar domain but do not belong to the original training data, AdaRound achieves competitive performance. Here, we observe a less than $0.2\%$ degradation on average.
It is worthwhile to note that both Pascal VOC and MS COCO only contain a small subset of the classes from Imagenet, implying that the optimization data for AdaRound does not need to be fully representative of the original training set.

\begin{figure}[t]
    \includegraphics[width=8cm, trim={.2cm .2cm .2cm .2cm}, clip]{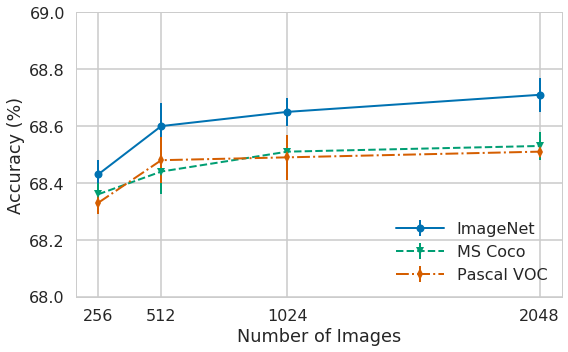}\vspace{-0.2cm}
    \centering
    \caption{The effect on ImageNet validation accuracy when using different number of images belonging to different datasets for AdaRound optimization.}
\label{fig:datasets}
\end{figure}

\begin{table*}[t]
    \centering
    \begin{tabular}{ l r r r r r}
        \toprule
         Optimization           & \#bits W/A    & Resnet18      & Resnet50  & InceptionV3   & MobilenetV2 \\\midrule
         Full precision         & 32/32         & 69.68         & 76.07     & 77.40         &  71.72 \\
         DFQ \citep{dfq}        & 8/8           &  69.7         &     -     &  -            & 71.2  \\
         \midrule
         Nearest                & 4/32          & 23.99         & 35.60     & 1.67          & 8.09 \\
         OMSE+opt\citep{huaweiquant} & 4$^*$/32    & 67.12      & 74.67     & 73.66         & - \\
         OCS \citep{OCS}        & 4/32          &  -            & 66.2      & 4.8           & - \\
         AdaRound               & 4/32          & \textbf{68.71$\pm$0.06} &  \textbf{75.23$\pm$0.04} & \textbf{75.76$\pm$0.09} &  \textbf{69.78$\pm$0.05}$^{\dagger}$  \\
         \midrule
         DFQ (our impl.)        & 4/8           & 38.98         & 52.84     & -          & 46.57 \\
         Bias corr \citep{bannerposttraining}
                                & 4$^*$/8          & 67.4          & 74.8      & 59.5      & - \\
         AdaRound w/ act quant   & 4/8           & \textbf{68.55$\pm$0.01} &  \textbf{75.01$\pm$0.05} &  \textbf{75.72$\pm$0.09}         &   \textbf{69.25$\pm$0.06}$^{\dagger}$ \\
         \bottomrule
    \end{tabular}\vspace{-0.1cm}
    \caption{Comparison among different post-training quantization strategies in the literature. We report results for various models in terms of ImageNet validation accuracy (\%). *Uses per-channel quantization. $^{\dagger}$Using CLE \cite{dfq} as preprocessing.}
    \label{tbl:exp07_lit}\vspace{-0.1cm}
\end{table*}

\subsection{Literature comparison}\label{sec:comp_literature}

\paragraph{Comparison to bias correction}
Several recent papers have addressed a specific symptom of the problem we describe with rounding-to-nearest \cite{bannerposttraining, biaswithbias, dfq}. These works observe that quantizing weights often changes the expected value of the output of the layer, i.e., $\eop{\mat{W}\vec{x}} \neq \mathbb{E}[\matq{W}\vec{x}]$. In order to counteract this, these papers adjust the bias terms for the preactivations by adding $\eop{\mat{W}\vec{x}} - \mathbb{E}[\matq{W}\vec{x}]$. This ``bias correction" can be viewed as another approach to minimize the same MSE loss as AdaRound \eqref{eq:subhessopt2}, but by adjusting the bias terms as
\begin{align}
 \eop{\mat{W}\vec{x}} - \mathbb{E}[\matq{W}\vec{x}] & = \argmin\limits_{\vecq{b}} \eop{\norm*{\mat{W}\vec{x} - \left(\matq{W}\vec{x} + \vecq{b}\right)}^2_F}.
\end{align}
Our method solves this same problem, but in a better way. In Table \ref{tbl:comp_bias_corr} we compare empirical bias correction from \citet{dfq} to AdaRound, under the exact same experimental setup, on ResNet18. While bias correction improves performance over vanilla quantization without bias correction, we see that for 4 bits it only achieves $38.87\%$ accuracy, where AdaRound recovers accuracy to $68.60\%$.

\begin{table}[t]
    \centering
    \begin{tabular}{ l r }
        \toprule
         Rounding           & Acc(\%)     \\\midrule
         Nearest            & 23.99    \\  
         Bias correction    & 38.87    \\  
         AdaRound           & \textbf{68.60$\pm$0.09}    \\
         \bottomrule 
    \end{tabular}\vspace{-0.1cm}
    \caption{Comparison between AdaRound and empirical bias correction, which also counteracts a symptom of the quantization error introduced by rounding to nearest. We report ImageNet validation accuracy (\%) for Resnet18.}
    \label{tbl:comp_bias_corr}\vspace{-0.1cm}
\end{table}

\paragraph{ImageNet}
In Table \ref{tbl:exp07_lit}, we compare AdaRound to several recent post-training quantization methods. 
We use the same experimental setup as described earlier, with the exception of optimizing AdaRound with 2048 images for 20k iterations. 
For both Resnet18 and Resnet50, AdaRound is within 1\% of the FP32 accuracy for 4-bit weight quantization and outperforms all competing methods, even though some rely on the more favorable per-channel quantization and do not quantize the first and the last layer. 
Similarly, on the more challenging networks, InceptionV3 and MobilenetV2, AdaRound stays within 2\% of the original accuracy and outperforms any competing method.

To be able to compare to methods that also do activation quantization, we report results of AdaRound with all activation tensors quantized to 8 bits. For this scenario, we quantized the activations to 8 bits and set the scaling factor for the activation quantizers based on the minimum and maximum activations observed. 
We notice that activation quantization, in most cases, does not significantly harm the validation accuracy.
AdaRound again outperforms the competing methods such as DFQ~\citep{dfq} and bias correction~\citep{bannerposttraining}.


\paragraph{Semantic segmentation}
To demonstrate the wider applicability of AdaRound, we apply it to DeeplabV3+ \cite{chen2018deeplab} evaluated on Pascal VOC \cite{everingham2015pascal}. Since the input images here are significantly bigger, we only use 512 images to optimize AdaRound. All other aspects of the experimental setup stay the same. 
To the best of our knowledge, there are no other post-training quantization methods doing 4-bit quantization for semantic segmentation. DFQ works well for 8 bits, however performance drastically drops when going down to 4-bit weight quantization. AdaRound still performs well for 4 bits and has only a 2\% performance decrease for 4-bit weights and 8-bit activations quantization.

\begin{table}[t]
    \centering
    \begin{tabular}{ l r r }
        \toprule
         Optimization          & \#bits W/A      & mIOU \\\midrule
         Full precision   & 32/32 & 72.94  \\
         DFQ \citep{dfq}   & 8/8   &  72.33 \\
         Nearest      &  4/8   &  6.09 \\
         DFQ (our impl.)   & 4/8   &  14.45 \\
         \midrule
         AdaRound    & 4/32   &  70.89$\pm$0.33 \\ 
         AdaRound w/ act quant    & 4/8   & 70.86$\pm$0.37  \\ 
         \bottomrule 
    \end{tabular}\vspace{-0.1cm}
    \caption{Comparison among different post-training quantization strategies, in terms of Mean Intersection Over Union (mIOU) for DeeplabV3+ (MobileNetV2 backend) on Pascal VOC.}
    \label{tbl:semseg}\vspace{-0.1cm}
\end{table}

\section{Conclusion}\label{sec:conclusion}
In this paper we proposed AdaRound, a new rounding method for post-training quantization of neural network weights. AdaRound improves significantly over rounding-to-nearest, which has poor performance for lower bit widths. We framed and analyzed the rounding problem theoretically and by making appropriate approximations 
we arrive at a practical method. AdaRound is computationally fast, uses only a small number of unlabeled data examples, does not need end-to-end fine-tuning, and can be applied to any neural network that has convolutional or fully-connected layers without any restriction. 
AdaRound establishes a new state-of-the-art for post-training  weight quantization with significant gains. It can push networks like Resnet18 and Resnet50 to 4-bit weights while keeping the accuracy drop within 1\%.

\FloatBarrier










\bibliography{dirty}

\begin{thebibliography}{39}
\providecommand{\natexlab}[1]{#1}
\providecommand{\url}[1]{\texttt{#1}}
\expandafter\ifx\csname urlstyle\endcsname\relax
  \providecommand{\doi}[1]{doi: #1}\else
  \providecommand{\doi}{doi: \begingroup \urlstyle{rm}\Url}\fi

\bibitem[Banner et~al.(2019)Banner, Nahshan, and Soudry]{bannerposttraining}
Banner, R., Nahshan, Y., and Soudry, D.
\newblock Post training 4-bit quantization of convolutional networks for
  rapid-deployment.
\newblock \emph{Neural Information Processing Systems (NeuRIPS)}, 2019.

\bibitem[Bengio et~al.(2013)Bengio, L{\'e}onard, and
  Courville]{bengio2013estimating}
Bengio, Y., L{\'e}onard, N., and Courville, A.
\newblock Estimating or propagating gradients through stochastic neurons for
  conditional computation.
\newblock \emph{arXiv preprint arXiv:1308.3432}, 2013.

\bibitem[Botev et~al.(2017)Botev, Ritter, and Barber]{gauss-newton2017}
Botev, A., Ritter, H., and Barber, D.
\newblock Practical gauss-newton optimisation for deep learning.
\newblock \emph{International Conference on Machine Learning (ICML)}, 2017.

\bibitem[Cai et~al.(2020)Cai, Yao, Dong, Gholami, Mahoney, and
  Keutzer]{zeroshotquant}
Cai, Y., Yao, Z., Dong, Z., Gholami, A., Mahoney, M.~W., and Keutzer, K.
\newblock Zeroq: {A} novel zero shot quantization framework.
\newblock \emph{arXiv preprint arXiv:2001.00281}, 2020.

\bibitem[Chen et~al.(2018)Chen, Zhu, Papandreou, Schroff, and
  Adam]{chen2018deeplab}
Chen, L.-C., Zhu, Y., Papandreou, G., Schroff, F., and Adam, H.
\newblock Encoder-decoder with atrous separable convolution for semantic image
  segmentation.
\newblock \emph{The European Conference on Computer Vision (ECCV)}, 2018.

\bibitem[Choi et~al.(2018)Choi, Wang, Venkataramani, Chuang, Srinivasan, and
  Gopalakrishnan]{pact2018}
Choi, J., Wang, Z., Venkataramani, S., Chuang, P.~I., Srinivasan, V., and
  Gopalakrishnan, K.
\newblock {PACT:} parameterized clipping activation for quantized neural
  networks.
\newblock \emph{arXiv preprint arxiv:805.06085}, 2018.

\bibitem[Choukroun et~al.(2019)Choukroun, Kravchik, and Kisilev]{huaweiquant}
Choukroun, Y., Kravchik, E., and Kisilev, P.
\newblock Low-bit quantization of neural networks for efficient inference.
\newblock \emph{International Conference on Computer Vision (ICCV)}, 2019.

\bibitem[Dong et~al.(2019)Dong, Yao, Gholami, Mahoney, and Keutzer]{hawq}
Dong, Z., Yao, Z., Gholami, A., Mahoney, M.~W., and Keutzer, K.
\newblock {HAWQ:} hessian aware quantization of neural networks with
  mixed-precision.
\newblock \emph{International Conference on Computer Vision (ICCV)}, 2019.

\bibitem[Esser et~al.(2020)Esser, McKinstry, Bablani, Appuswamy, and
  Modha]{lsq}
Esser, S.~K., McKinstry, J.~L., Bablani, D., Appuswamy, R., and Modha, D.~S.
\newblock Learned step size quantization.
\newblock \emph{International Conference on Learning Representations (ICLR)},
  2020.

\bibitem[Everingham et~al.(2015)Everingham, Eslami, {Van Gool}, Williams, Winn,
  and Zisserman]{everingham2015pascal}
Everingham, M., Eslami, S., {Van Gool}, L., Williams, C., Winn, J., and
  Zisserman, A.
\newblock The pascal visual object classes challenge: A retrospective.
\newblock \emph{International Journal of Computer Vision}, 111\penalty0
  (1):\penalty0 98--136, 1 2015.

\bibitem[Finkelstein et~al.(2019)Finkelstein, Almog, and Grobman]{biaswithbias}
Finkelstein, A., Almog, U., and Grobman, M.
\newblock Fighting quantization bias with bias.
\newblock \emph{arXiv preprint arxiv:1906.03193}, 2019.

\bibitem[Guo(2018)]{guosurvey}
Guo, Y.
\newblock A survey on methods and theories of quantized neural networks.
\newblock \emph{arXiv preprint: arxiv:1808.04752}, 2018.

\bibitem[Gupta et~al.(2015)Gupta, Agrawal, Gopalakrishnan, and
  Narayanan]{Gupta2015}
Gupta, S., Agrawal, A., Gopalakrishnan, K., and Narayanan, P.
\newblock Deep learning with limited numerical precision.
\newblock \emph{International Conference on Machine Learning, {ICML}}, 2015.

\bibitem[{Hammerstrom}(1990)]{hammerstrom}
{Hammerstrom}, D.
\newblock A vlsi architecture for high-performance, low-cost, on-chip learning.
\newblock \emph{International Joint Conference on Neural Networks (IJCNN)},
  1990.

\bibitem[He et~al.(2016)He, Zhang, Ren, and Sun]{heresidual}
He, K., Zhang, X., Ren, S., and Sun, J.
\newblock Deep residual learning for image recognition.
\newblock \emph{Conference on Computer Vision and Pattern Recognition, {CVPR}},
  2016.

\bibitem[He et~al.(2017)He, Zhang, and Sun]{he2017}
He, Y., Zhang, X., and Sun, J.
\newblock Channel pruning for accelerating very deep neural networks.
\newblock \emph{International Conference on Computer Vision (ICCV)}, 2017.

\bibitem[{Hoehfeld} \& {Fahlman}(1992){Hoehfeld} and
  {Fahlman}]{oldstochasticrounding}
{Hoehfeld}, M. and {Fahlman}, S.~E.
\newblock Learning with limited numerical precision using the
  cascade-correlation algorithm.
\newblock \emph{IEEE Transactions on Neural Networks}, 3\penalty0 (4):\penalty0
  602--611, 1992.

\bibitem[Holi \& Hwang(1993)Holi and Hwang]{holihwang}
Holi, J.~L. and Hwang, J.~N.
\newblock Finite precision error analysis of neural network hardware
  implementations.
\newblock \emph{IEEE Trans. Comput.}, 42\penalty0 (3):\penalty0 281–290,
  1993.

\bibitem[Hopfield \& Tank(1985)Hopfield and Tank]{Hopfield1985}
Hopfield, J.~J. and Tank, D.~W.
\newblock ``neural'' computation of decisions in optimization problems.
\newblock \emph{Biological Cybernetics}, 52\penalty0 (3):\penalty0 141--152,
  1985.

\bibitem[Jacob et~al.(2018)Jacob, Kligys, Chen, Zhu, Tang, Howard, Adam, and
  Kalenichenko]{jacob2018cvpr}
Jacob, B., Kligys, S., Chen, B., Zhu, M., Tang, M., Howard, A., Adam, H., and
  Kalenichenko, D.
\newblock Quantization and training of neural networks for efficient
  integer-arithmetic-only inference.
\newblock \emph{Conference on Computer Vision and Pattern Recognition (CVPR)},
  2018.

\bibitem[Jain et~al.(2019)Jain, Gural, Wu, and Dick]{xilinxquant}
Jain, S.~R., Gural, A., Wu, M., and Dick, C.
\newblock Trained uniform quantization for accurate and efficient neural
  network inference on fixed-point hardware.
\newblock \emph{arxiv preprint arxiv:1903.08066}, 2019.

\bibitem[Kim et~al.(2019)Kim, Bhalgat, Lee, Patel, and Kwak]{jangho}
Kim, J., Bhalgat, Y., Lee, J., Patel, C., and Kwak, N.
\newblock {QKD:} quantization-aware knowledge distillation.
\newblock \emph{arxiv preprint arxiv:1911.12491}, 2019.

\bibitem[Kingma \& Ba(2015)Kingma and Ba]{kingma2014adam}
Kingma, D.~P. and Ba, J.
\newblock Adam: A method for stochastic optimization.
\newblock \emph{International Conference for Learning Representations (ICLR)},
  2015.

\bibitem[Kochenberger et~al.(2014)Kochenberger, Hao, Glover, Lewis, L{\"u},
  Wang, and Wang]{Kochenberger2014}
Kochenberger, G., Hao, J.-K., Glover, F., Lewis, M., L{\"u}, Z., Wang, H., and
  Wang, Y.
\newblock The unconstrained binary quadratic programming problem: a survey.
\newblock \emph{Journal of Combinatorial Optimization}, 28\penalty0
  (1):\penalty0 58--81, Jul 2014.

\bibitem[{Krishnamoorthi}(2018)]{krishnamoorthi}
{Krishnamoorthi}, R.
\newblock {Quantizing deep convolutional networks for efficient inference: A
  whitepaper}.
\newblock \emph{arXiv preprint arXiv:1806.08342}, 2018.

\bibitem[Lin et~al.(2016)Lin, Talathi, and Annapureddy]{lin2016icml}
Lin, D.~D., Talathi, S.~S., and Annapureddy, V.~S.
\newblock Fixed point quantization of deep convolutional networks.
\newblock In \emph{International Conference on Machine Learning}, 2016.

\bibitem[Louizos et~al.(2018)Louizos, Welling, and Kingma]{louizos2018learning}
Louizos, C., Welling, M., and Kingma, D.~P.
\newblock Learning sparse neural networks through $l_0$ regularization.
\newblock \emph{International Conference on Learning Representations (ICLR)},
  2018.

\bibitem[Louizos et~al.(2019)Louizos, Reisser, Blankevoort, Gavves, and
  Welling]{louizos2018relaxed}
Louizos, C., Reisser, M., Blankevoort, T., Gavves, E., and Welling, M.
\newblock Relaxed quantization for discretized neural networks.
\newblock In \emph{International Conference on Learning Representations
  (ICLR)}, 2019.

\bibitem[Mishra \& Marr(2017)Mishra and Marr]{apprentice}
Mishra, A.~K. and Marr, D.
\newblock Apprentice: Using knowledge distillation techniques to improve
  low-precision network accuracy.
\newblock \emph{arXiv preprint arxiv:1711.05852}, 2017.

\bibitem[Nagel et~al.(2019)Nagel, van Baalen, Blankevoort, and Welling]{dfq}
Nagel, M., van Baalen, M., Blankevoort, T., and Welling, M.
\newblock Data-free quantization through weight equalization and bias
  correction.
\newblock \emph{International Conference on Computer Vision (ICCV)}, 2019.

\bibitem[Paszke et~al.(2019)Paszke, Gross, Massa, Lerer, Bradbury, Chanan,
  Killeen, Lin, Gimelshein, Antiga, Desmaison, Kopf, Yang, DeVito, Raison,
  Tejani, Chilamkurthy, Steiner, Fang, Bai, and Chintala]{pytorch}
Paszke, A., Gross, S., Massa, F., Lerer, A., Bradbury, J., Chanan, G., Killeen,
  T., Lin, Z., Gimelshein, N., Antiga, L., Desmaison, A., Kopf, A., Yang, E.,
  DeVito, Z., Raison, M., Tejani, A., Chilamkurthy, S., Steiner, B., Fang, L.,
  Bai, J., and Chintala, S.
\newblock Pytorch: An imperative style, high-performance deep learning library.
\newblock In \emph{Neural Information Processing Systems (NeuRIPS)}. 2019.

\bibitem[Rubinstein(1999)]{Rubinstein1999}
Rubinstein, R.
\newblock The cross-entropy method for combinatorial and continuous
  optimization.
\newblock \emph{Methodology And Computing In Applied Probability}, 1\penalty0
  (2):\penalty0 127--190, Sep 1999.

\bibitem[Russakovsky et~al.(2015)Russakovsky, Deng, Su, Krause, Satheesh, Ma,
  Huang, Karpathy, Khosla, Bernstein, Berg, and Fei-Fei]{imagenet}
Russakovsky, O., Deng, J., Su, H., Krause, J., Satheesh, S., Ma, S., Huang, Z.,
  Karpathy, A., Khosla, A., Bernstein, M., Berg, A.~C., and Fei-Fei, L.
\newblock {ImageNet Large Scale Visual Recognition Challenge}.
\newblock \emph{International Journal of Computer Vision (IJCV)}, 115\penalty0
  (3):\penalty0 211--252, 2015.

\bibitem[Smith et~al.()Smith, Palaniswami, and Krishnamoorthy]{smithhopfield}
Smith, K.~A., Palaniswami, M., and Krishnamoorthy, M.
\newblock Neural techniques for combinatorial optimization with applications.
\newblock \emph{{IEEE} Trans. Neural Networks}, 9\penalty0 (6):\penalty0
  1301--1318.

\bibitem[Stock et~al.(2020)Stock, Joulin, Gribonval, Graham, and
  Jégou]{BGD2020}
Stock, P., Joulin, A., Gribonval, R., Graham, B., and Jégou, H.
\newblock And the bit goes down: Revisiting the quantization of neural
  networks.
\newblock In \emph{International Conference on Learning Representations}, 2020.

\bibitem[Uhlich et~al.(2020)Uhlich, Mauch, Yoshiyama, Cardinaux, Garc{\'{\i}}a,
  Tiedemann, Kemp, and Nakamura]{differentiablequantization}
Uhlich, S., Mauch, L., Yoshiyama, K., Cardinaux, F., Garc{\'{\i}}a, J.~A.,
  Tiedemann, S., Kemp, T., and Nakamura, A.
\newblock Mixed precision dnns: All you need is a good parametrization.
\newblock \emph{International Conference on Learning Representations (ICLR)},
  2020.

\bibitem[{Wang} et~al.(2018){Wang}, {Hu}, {Zhang}, {Zhang}, {Liu}, and
  {Cheng}]{TSQ2018}
{Wang}, P., {Hu}, Q., {Zhang}, Y., {Zhang}, C., {Liu}, Y., and {Cheng}, J.
\newblock Two-step quantization for low-bit neural networks.
\newblock \emph{Conference on Computer Vision and Pattern Recognition (CVPR)},
  pp.\  4376--4384, 2018.

\bibitem[Zhang et~al.(2016)Zhang, Zou, He, and Sun]{zhang2015}
Zhang, X., Zou, J., He, K., and Sun, J.
\newblock Accelerating very deep convolutional networks for classification and
  detection.
\newblock \emph{{IEEE} Trans. Pattern Anal. Mach. Intell.}, 38\penalty0
  (10):\penalty0 1943--1955, 2016.

\bibitem[Zhao et~al.(2019)Zhao, Hu, Dotzel, Sa, and Zhang]{OCS}
Zhao, R., Hu, Y., Dotzel, J., Sa, C.~D., and Zhang, Z.
\newblock Improving neural network quantization without retraining using
  outlier channel splitting.
\newblock \emph{International Conference on Machine Learning, {ICML}}, 2019.

\end{thebibliography}
\bibliographystyle{icml2020}
\fi

\ifappendix
\onecolumn
\icmltitle{Up or Down? Adaptive Rounding for Post-Training Quantization}
\appendix
\section{Comparison among QUBO solvers}\label{app:qubo_solvers}
We compared optimizing task loss Hessian using the cross-entropy method vs QUBO solver from the publicly available package \textit{qbsolv}\footnote{\url{https://docs.ocean.dwavesys.com/projects/qbsolv/}}. We chose this qbsolv QUBO solver for comparison due to its ease of use for our needs as well its free availability for any researcher to reproduce our work.  Table~\ref{tbl:qubosolvers} presents the comparison between the two solvers. We see that cross-entropy method significantly outperforms the \textit{qbsolv} QUBO solver. Furthermore the \textit{qbsolv} QUBO solver has worse performance than rounding-to-nearest. We believe this is mainly due to the reason that the API does not allow us to provide a smart initialization (as we do for cross-entropy method). The performance of random rounding choices is significantly worse, on average, when compared to the rounding choices in the neighbourhood of rounding-to-nearest. Hence this initialization can provide a significant advantage in finding a better local minimum in this large problem space. We did not conduct an extensive search for better QUBO solvers as our own implementation of the cross-entropy method provided very good results with very little tweaking and allowed us to exploit GPU and memory resources more efficiently. Furthermore the choice of QUBO solver does not impact our final method AdaRound while clearly showing the gains that we can exploit via optimized rounding. 

\begin{table}[bth]
    \centering
    \begin{tabular}{ l r }
        \toprule
         Rounding           & First layer     \\\midrule
         Nearest            & 52.29    \\  
         Cross-entropy Method    & 68.62$\pm$0.17    \\  
         QUBO solver (qbsolv)           & 41.98$\pm$3.04    \\
         \bottomrule 
    \end{tabular}\vspace{-0.1cm}
    \caption{Comparison between the cross-entropy method vs \textit{qbsolv} QUBO solver. Only the first layer of Resnet18 is quantized to 4-bits and the results are reported in terms of ImageNet validation accuracy.}
    \label{tbl:qubosolvers}\vspace{-0.1cm}
\end{table}

\section{From Taylor expansion to local loss (conv. layer)}\label{app:convlayer}
For a convolutional layer, defined as $\veci{z}{\ell} = \mati{W}{\ell} \ast \veci{x}{\ell-1}$, we have 
\begin{align}
     \frac{\partial \tlossb}{\partial \mati{W}{\ell}_{h_1,w_1,c^{i}_1,c^{o}_1}} &= \sum\limits_{i,j} \frac{\partial \veci{z}{\ell}_{i,j,c^{o}_1}}{\partial \mati{W}{\ell}_{h_1,w_1,c^{i}_1,c^{o}_1}} \cdot \frac{\partial \tlossb}{\partial \veci{z}{\ell}_{i,j,c^{o}_1}} \\
     &= \sum\limits_{i,j} \frac{\partial \tlossb}{\partial \veci{z}{\ell}_{i,j,c^{o}_{1}}} \cdot \veci{x}{\ell-1}_{i+h_1,j+w_1,c^{i}_1},\label{eq:fdiffconv}
\end{align}
where $h_1$ and $w_1$ denote the spatial dimensions, $c^i_1$ denotes input channel dimension and $c^o_1$ denotes output channel dimension. Additionally, we have assumed appropriate zero padding of $\veci{x}{\ell-1}$. Differentiating \eqref{eq:fdiffconv} once again (possibly w.r.t. a different weight in the same layer), we get
\begin{align}
     \frac{\partial^2 \tlossb}{\partial \mati{W}{\ell}_{h_1,w_1,c^{i}_1,c^{o}_1}  \partial \veci{W}{\ell}_{h_2,w_2,c^{i}_2,c^{o}_2}}
     &= \sum\limits_{i,j} \sum\limits_{k,m}  \veci{x}{\ell-1}_{i+h_1,j+w_1,c^{i}_1} \veci{x}{\ell-1}_{k+h_2,m+w_2,c^{i}_2} \cdot \frac{\partial^2 \tlossb}{\partial \veci{z}{\ell}_{i,j,c^{o}_1} \partial\veci{z}{\ell}_{k,m,c^{o}_2}}. \label{eq:secdiffconv}
\end{align}
In order to transform the Hessian QUBO optimization problem to a local loss based per-layer optimization problem, we assume that $\nabla^2_{\veci{z}{\ell}} \tlossb$ is a diagonal matrix that is independent of the data samples $\left(\vec{x},\vec{y}\right)$, i.e., 
\begin{equation}
  \frac{\partial^2 \tlossb}{\partial \veci{z}{\ell}_{i,j,c^{o}_1} \partial\veci{z}{\ell}_{k,m,c^{o}_2}}=\begin{cases}
    \ct{c}_{c^{o}_1}, & \text{if $i=k,j=m,c^{o}_1=c^{o}_2$}\\
    0, & \text{otherwise}.
  \end{cases}\label{eq:convassumption}
\end{equation}
This assumption reduces \eqref{eq:secdiffconv} to 
\begin{equation}\label{eq:diagdiffconv}
  \frac{\partial^2 \tlossb}{\partial \mati{W}{\ell}_{h_1,w_1,c^{i}_1,c^{o}_1}  \partial \mati{W}{\ell}_{h_2,w_2,c^{i}_2,c^{o}_2}}=\begin{cases}
     \ct{c}_{c^{o}_1} \sum\limits_{i,j}\veci{x}{\ell-1}_{i+h_1,j+w_1,c^{i}_1} \veci{x}{\ell-1}_{i+h_2,j+w_2,c^{i}_2}, & \text{if $c^{o}_1=c^{o}_2$}\\
    0, & \text{otherwise}.
  \end{cases}
\end{equation}
Under the assumptions in \eqref{eq:convassumption} there are no interactions between weights in the same layer that affect two different output filters $\left(c^{o}_1 \neq c^{o}_2\right)$. We then reformulate the Hessian QUBO optimization
\begin{align}
    &\eop{\dw^{(\ell), T}\mati{H}{\veci{w}{\ell}} \dw^{(\ell)}}  \\
    &\overset{(a)}{=} \eop{\sum\limits_{c^{o}}\ct{c}_{c^{o}}\sum\limits_{h_1,w_1,c^{i}_1} \sum\limits_{h_2,w_2,c^{i}_2} \sum\limits_{i,j}  \Delta \mati{W}{\ell}_{h_1,w_1,c^{i}_1,c^{o}}   \Delta \mati{W}{\ell}_{h_2,w_2,c^{i}_2,c^{o}} \veci{x}{\ell-1}_{i+h_1,j+w_1,c^{i}_1} \veci{x}{\ell-1}_{i+h_2,j+w_2,c^{i}_2}} \\
    &= \eop{\sum\limits_{c^{o}}\ct{c}_{c^{o}} \sum\limits_{i,j} \left(\sum\limits_{h,w,c^{i}} \Delta \mati{W}{\ell}_{h,w,c^{i},c^{o}} \veci{x}{\ell-1}_{i+h,j+w,c^{i}}\right)^2} \\
    &= \eop{\sum\limits_{c^{o}}\ct{c}_{c^{o}} \norm*{\Delta \mati{W}{\ell}_{:,:,:,c^{o}} \ast \veci{x}{\ell-1}}^2_F}, \label{eq:convMSE}
\end{align}
where (a) follows from the assumption in \eqref{eq:convassumption}. Hence the Hessian optimization problem, under the assumptions in \eqref{eq:convassumption}, is the same as MSE optimization for the output feature map. Furthermore, it breaks down to an optimization problem for each individual output channel separately (each element in the summation in \eqref{eq:convMSE} is independent of the other elements in the summation for optimization purposes as they involve disjoint sets of variables). 
\begin{align}
    \argmin_{\dw^{(\ell)}} \quad \eop{\dw^{(\ell), T}\mati{H}{\veci{w}{\ell}} \dw^{(\ell)}} &= \argmin_{\Delta \mati{W}{\ell}} \eop{\norm*{\Delta \mati{W}{\ell} \ast \veci{x}{\ell-1}}^2_F} \\
    &= \argmin_{\Delta \mati{W}{\ell}_{:,:,:,c^{o}}}  \eop{\norm*{\Delta \mati{W}{\ell}_{:,:,:,c^{o}} \ast \veci{x}{\ell-1}}^2_F} \qquad \forall{c^{o}}.
\end{align}

\fi
%



\end{document}